\documentclass[10pt,twocolumn,letterpaper]{article}

\usepackage[pagenumbers]{iccv}      
\usepackage{colortbl}
\usepackage{multirow}
\usepackage{multicol}

\definecolor{iccvblue}{rgb}{0.21,0.49,0.74}
\usepackage[pagebackref,breaklinks,colorlinks,allcolors=iccvblue]{hyperref}

\def\paperID{*****} 
\def\confName{ICCV}
\def\confYear{2025}

\title{Step by Step Network}

\author{
Dongchen Han$^1$\hspace{1mm}
Tianzhu Ye$^{1}$\hspace{1mm}
Zhuofan Xia$^{1}$\hspace{1mm}
Kaiyi Chen$^{1}$\hspace{1mm}
Yulin Wang$^{1}$\hspace{1mm}
Hanting Chen$^{2}$\hspace{1mm}
Gao Huang$^{1}\thanks{Corresponding Author.}$ \\
{\small $^1$ Tsinghua University\hspace{6mm}
$^2$ Huawei Noah’s Ark Lab}
}

\begin{document}
\maketitle
\begin{abstract}

Scaling up network depth is a fundamental pursuit in neural architecture design, as theory suggests that deeper models offer exponentially greater capability.
Benefiting from the residual connections, modern neural networks can scale up to more than one hundred layers and enjoy wide success. 
However, as networks continue to deepen, current architectures often struggle to realize their theoretical capacity improvements, calling for more advanced designs to further unleash the potential of deeper networks.
In this paper, we identify two key barriers that obstruct residual models from scaling deeper: shortcut degradation and limited width.
Shortcut degradation hinders deep-layer learning, while the inherent depth-width trade-off imposes limited width.
To mitigate these issues, we propose a generalized residual architecture dubbed \textbf{Step by Step Network (StepsNet)} to bridge the gap between theoretical potential and practical performance of deep models.
Specifically, we separate features along the channel dimension and let the model learn progressively via stacking blocks with increasing width. 
The resulting method mitigates the two identified problems and serves as a versatile macro design applicable to various models. 
Extensive experiments show that our method consistently outperforms residual models across diverse tasks, including image classification, object detection, semantic segmentation, and language modeling. 
These results position StepsNet as a superior generalization of the widely adopted residual architecture.

\end{abstract}    
\section{Introduction}



In recent years, great success in computer vision has been witnessed through scaling up network depth. Early theoretical works~\cite{shallow_vs_deep, depth_power, expressive_power, exponential_expressivity} suggest that deeper neural networks can capture exponentially more informative representations than shallower ones. 
This theoretical insight propelled one of the most magnificent works, ResNet~\cite{resnet}, whose residual architecture allows networks with more than one hundred layers to be optimized effectively. Benefited from increased depth, residual models can effectively learn from large scale of data and enjoy substantial improvements, driving advances in the field of computer vision~\cite{vit, deit, convnext, dit} and various other scenarios~\cite{attention, clip}.

\begin{figure}[t]
    \centering
    \includegraphics[width=\linewidth]{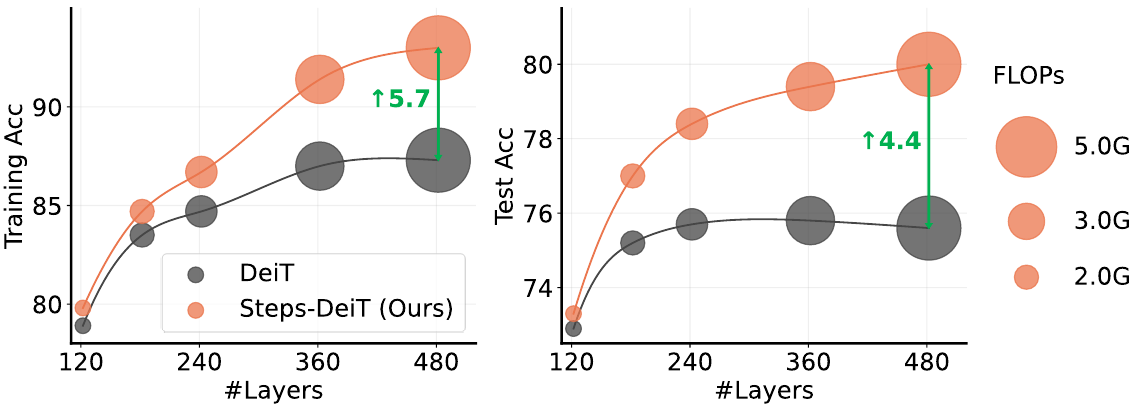}
    \vskip -0.2cm
    \caption{Training accuracy (left) and test accuracy (right) on ImageNet-1K as we gradually increase model depth while keeping the width fixed. The residual architecture DeiT does not deliver satisfactory results as depth increases. In contrast, our method enables the model to leverage increased depth more effectively, achieving much higher results on both training and test sets. Notably, the DeiT models used in this pilot study are \textit{not standard DeiT-T/S/B}. Please refer to \cref{sec:exp_deeper} for experiment details. }
    \vskip -0.4cm
    \label{fig:deeper}
\end{figure}

However, have we fully revealed the potential of network depth? Maybe not. Despite the theoretical power of very deep networks, modern CNNs~\cite{mobilenetv2, convnext} and Vision Transformers~\cite{vit, deit, swin} generally saturate their optimal performance at some relatively modest depths around one hundred layers.
As one simply increase depth further, these models remain trainable but often struggle to deliver enough improvements. 
Notably, the core reason behind this phenomenon is \textit{not} simple overfitting~\cite{deepvit, layerscale}.
A typical example is shown in \cref{fig:deeper}, where the deeper DeiT models 
do not demonstrate the theoretical gain in training accuracy, and thus the test results. This disparity from theory indicates that the widely successful residual architecture may still exist unresolved challenges at increased depth, which possibly hinders the real expressive capability of deeper networks. 
This motivates us to rethink the design of residual connections and come up with a fundamental research question: 

\textit{How should we further scale up the network depth effectively and unleash the great potential of deeper networks?}

In this paper, we start by reconsidering the residual architecture in deep models, offering both theoretical and empirical analyses to shed some light on this crucial question. Specifically, we analyze the key challenges of deep residual networks from two perspectives: shortcut degradation and limited width. Firstly, we reveal that the shortcuts in deep residual models can ``degrade'' during early training, which prevents deep layers from accessing input information and propagating back their gradients, resulting in optimization difficulties. Secondly, the inherent width-depth trade-off limits deep residual models to insufficient width. Under a practical computational budget, deeper residual models have to be restricted to thinner widths than their shallower counterparts, impairing their expressive capacity~\cite{universal_approx, min_width, min_width_2}.


Based on our analysis, we propose a generalized residual architecture, \textbf{Step by Step Network (StepsNet)}, as a promising remedy for these limitations. Instead of processing the input all at once, we split it into multiple parts along the channel dimension and progressively take each part of the information into consideration. This leads to a macro architecture composed of multiple residual networks of increasing widths, following a narrow-to-wide stacking strategy (see \cref{fig:stepsnet}). We demonstrate that our design improves the conventional residual architecture with easier optimization at large depth and mitigates its width-depth trade-off, thereby effectively overcoming the two identified limitations and further revealing the potential of deep networks. Empirical studies across diverse tasks are conducted to validate the effectiveness of StepsNet, including image classification, object detection, semantic segmentation and language modeling. The results, with an example shown in \cref{fig:deeper}, confirm that StepsNet benefits from increased depth and suggest it as an simple, effective and scalable generalization of the widely adopted residual architecture.


Our main contributions and takeaways are three-folded:
\begin{itemize}
    \item Through both theoretical and empirical analyses, we identify two key challenges of deep residual models: shortcut degradation and limited width.

    \item We present a simple, clean yet effective residual architecture named Step by Step Network (StepsNet), which mitigates these two challenges and supports the development of deeper models with enhanced expressive power. 
    StepsNet is independent of the micro designs and applicable to various residual models. 

    \item Extensive empirical evaluations across diverse tasks confirm that StepsNet enables models to benefit from increased depth, achieving improved results on conventional residual models without additional modifications.
\end{itemize}

\section{Related Works}

\noindent
\textbf{Theoretical power of network depth.}
Understanding the impact of depth is a fundamental research question in deep learning, with many studies offering theoretical insights. Early studies~\cite{shallow_vs_deep, depth_power} reveal that certain functions can be represented far more efficiently by deep networks than shallow ones. \cite{exponential_expressivity, expressive_power} further demonstrate that deep networks can represent exponentially more complex functions with increasing depth but not width. Other researches~\cite{number_of_linear, counting_linear_regions} show that the number of linear regions grows exponentially with network depth, but only polynomially with width. \cite{benefits_of_depth} proves that there exist deeper networks that can not be approximated by shallower ones without an exponential increase in size. Given the dramatic power of network depth, going deeper has long been the pursuit of neural networks, with extremely deep models being considered promising to address highly complex tasks.

\noindent
\textbf{Network architectures} are known to significantly impact the capacity and optimization of neural networks. ResNet~\cite{resnet} introduces residual connections, which greatly benefit network optimization and enable the development of deeper, more expressive models. Fractalnet~\cite{fractalnet} proposes the fractal architecture with paths of varying depths, matching ResNet’s performance without shortcuts. DenseNet~\cite{densenet} enhances information flow by using dense connections, achieving high parameter efficiency. Building on residual architectures, the Transformer model~\cite{attention} is proposed in natural language processing and quickly adopted in computer vision~\cite{vit, flatten, agent_attention, wang2025emulating}, achieving success in image classification~\cite{pvt, swin, inline, demystify_mamba}, semantic segmentation~\cite{maskformer, segformer}, object detection~\cite{detr,vitdet}, and multi-modal tasks~\cite{clip}.

\noindent
\textbf{Deep models.}
Motivated by the potential of very deep networks, many works are conducted to develop expressive deep models. For instance, ReZero~\cite{rezero} and LayerScale~\cite{layerscale} introduce learnable scale parameters in residual branches to improve convergence in deep residual models. FixUp~\cite{fixup} and DeepNet~\cite{deepnet} propose initialization strategies to stabilize deep model training. DeepViT~\cite{deepvit} addresses attention collapse in deep vision Transformers through re-attention. Normformer~\cite{normformer} improves transformer pretraining with extra normalization. Despite their elegant outcomes, these approaches still rely on conventional residual architectures, potentially limiting the capabilities of very deep networks. In this paper, we start with analyzing the challenges of deep residual models, and introduce a novel macro architecture named Step by Step Network to further unleash the power of depth. Our method serves as a strong alternative to residual architectures, compatible with various models, micro designs, and initialization techniques.

\begin{figure*}[t]
    \centering
    \includegraphics[width=\linewidth]{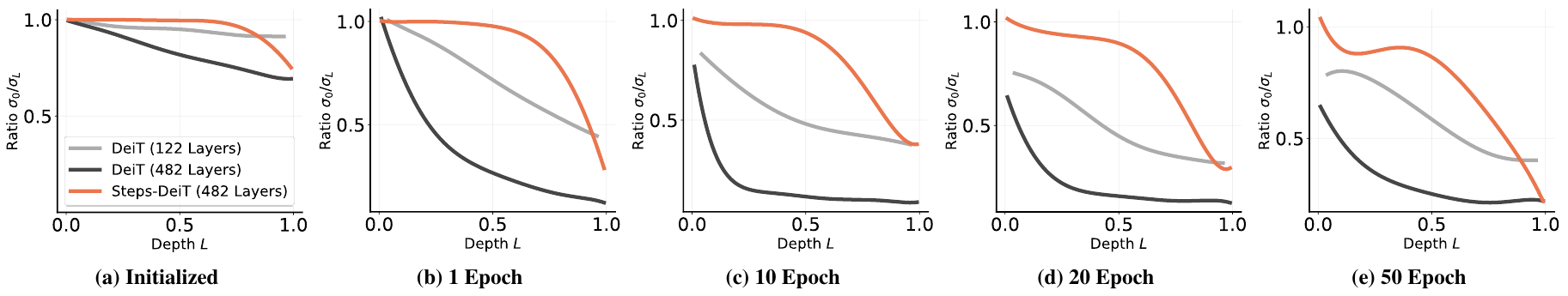}
    \vskip -0.2cm
    \caption{\textbf{The shortcut ratio $\gamma_l=\frac{\sigma_0}{\sigma_l}$ in DeiT and Steps-DeiT}, where $\sigma_0$ and $\sigma_l$ are the standard deviations of the input $z_0$ and feature after $l$ blocks $z_l$, respectively. The depth is normalized to $[0, 1]$, where 0 and 1 denote input and output. In a very deep residual model (more than 400 layers), the shortcut ratio $\frac{\sigma_0}{\sigma_l}$ approaches zero at early training stages, which prevents later residual blocks from obtaining input information and propagating its gradient back to the input, thus leading to optimization difficulties.}
    \label{fig:std}
    \vskip -0.2cm
\end{figure*}

\begin{figure}[t]
    \centering
    \includegraphics[width=0.9\linewidth]{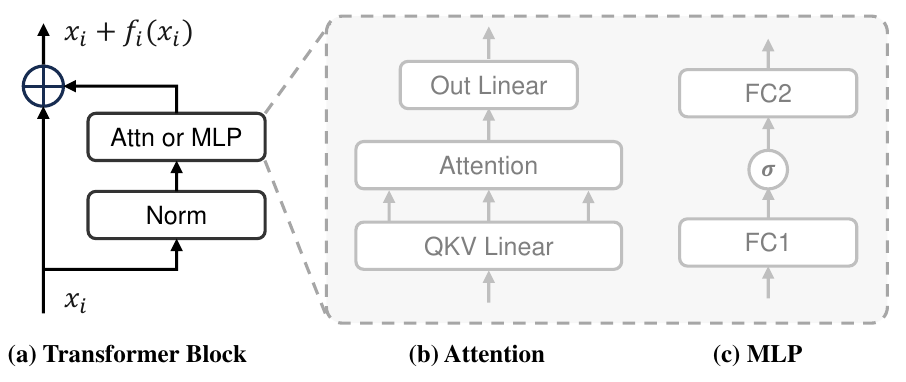}
    \vskip -0.2cm
    \caption{An illustration of a Transformer block to help understand the analyses in \cref{sec:challenges}. }
    \vskip -0.3cm
    \label{fig:transformer}
\end{figure}

\section{Challenges of Very Deep Residual Models}
\label{sec:challenges}

Theoretical studies indicate that the expressive power of neural networks grows exponentially with depth~\cite{expressive_power, exponential_expressivity}. Therefore, extremely deep networks are believed to have great potential for capturing intricate patterns and solving complex tasks. While the widely adopted residual architecture~\cite{resnet} allows models to benefit from over one hundred layers, further increasing depth often fails to yield expected returns. In this section, we dive into the challenges of very deep residual models, offering detailed analyses from two perspectives: shortcut degradation and limited width.

\subsection{Shortcut Degradation}

Residual architectures have been prominent in deep learning since the advent of ResNet~\cite{resnet}. They can be formulated as a sequence of blocks of the form:
\begin{equation}
\begin{split}
    z_l=z_{l-1}\!+\!\mathcal{R}_l(z_{l-1})=z_0\!+\!\!\sum_{i=1}^{l}\!\mathcal{R}_i(z_{i-1})\triangleq z_0\!+\!r_l,
\end{split}
\end{equation}
where $z_l$ is the representation after $l$ blocks, $\mathcal{R}_l$ denotes the $l$-th residual block, and $r_l=\sum_{i=1}^{l}\mathcal{R}_i(z_{i-1})$. 
Let's consider how $z_l$ is processed by the next block $\mathcal{R}_{l+1}$. Taking the Transformer architecture in \cref{fig:transformer} as an example, a normalization layer at the input of $\mathcal{R}_{l+1}$ transforms $z_l$ into:
\begin{equation}
\label{eq:std}
\begin{split}
    \hat{z_l}&=(z_0+r_l-\mu_l)/\sigma_l\\
    &=\frac{\sigma_0}{\sigma_l}\hat{z_0}+\frac{\sigma_r}{\sigma_l}\hat{r_l}+\frac{\mu_0+\mu_r-\mu_l}{\sigma_l}\\
    &=\frac{\sigma_0}{\sigma_l}\hat{z_0}+\frac{\sigma_r}{\sigma_l}\hat{r_l},
\end{split}
\end{equation}
where $\mu_0, \mu_l, \mu_r$, $\sigma_0, \sigma_l, \sigma_r$ and $\hat{z_0}, \hat{z_l}, \hat{r_l}$ are the means, standard deviations and normalized results of $z_0, z_l, r_l$, respectively. Note that $\mu_l\!=\!\mu_0\!+\!\mu_r$, as $z_l\!=\!z_0\!+\!r_l$. It is shown that the normalized $\hat{z_l}$ is a weighted combination of $\hat{z_0}$ and $\hat{r_l}$, weighted by $\frac{\sigma_0}{\sigma_l}$ and $\frac{\sigma_r}{\sigma_l}$. In early training stages, the $r_l\!=\!\sum_{i=1}^{l}\!\mathcal{R}_i(z_{i-1})$ contains minimal useful patterns due to randomly initialized $\mathcal{R}_i$ parameters. Consequently, $\hat{z_l}$ can be interpreted as a mixture of the input $\hat{z_0}$ and noisy residuals $\hat{r_l}$. We define the shortcut ratio $\gamma_l=\frac{\sigma_0}{\sigma_l}$ to quantify the preservation of the input shortcut up to depth $l$.

Previous studies demonstrate that feature variance $\sigma_l$ grows with network depth in residual architectures~\cite{start_train, fixup, skipinit}, thus causing $\gamma_l$ to decrease as $l$ increases. 
In \cref{fig:std}, we report $\gamma_l$ at various training epochs for the DeiT models presented in \cref{fig:deeper}. Consistent with our analysis, the 122-layer DeiT exhibits a decreasing $\gamma_l$ with depth, with an average value of around 0.5. In contrast, the 482-layer DeiT, which contains more residual terms, exhibits a rapidly diminishing shortcut ratio $\gamma_l$ that quickly approaches zero. Similar to the analyses in \cite{deepnet, rezero, skipinit}, we find such near-zero shortcut ratio $\gamma_l$ can lead to two critical issues in early training phases:
\begin{itemize}
    \item $\hat{z}_l =\frac{\sigma_0}{\sigma_l}\hat{z_0}+\frac{\sigma_r}{\sigma_l}\hat{r_l} \approx \frac{\sigma_r}{\sigma_l}\hat{r}_l$ becomes dominated by noisy residuals rather than the shortcut, preventing the block $\mathcal{R}_{l+1}$ from obtaining the input information $\hat{x}_0$.
    \item The gradient $\partial \hat{z}_l / \partial \hat{z}_0 = \sigma_0/\sigma_l \approx 0$ vanishes, hindering effective back propagation from $\mathcal{R}_{l+1}$ to the input.
\end{itemize}

These analyses suggest that while current residual architecture enables models with around one hundred layers to be trained effectively, the shortcuts in even deeper residual models can ``degrade'' during training, thus leading to optimization difficulties.
Notably, this challenge is not confined to the Transformer architecture, and similar analyses work for other residual models. The key observation is that very deep residual models add too much ``noise'' to the shortcut, and the proportion of input information diminishes with depth, resulting in both forward and backward issues.
Moreover, $z_0$ in our analysis may represent not only the raw input data but also features extracted by several early layers or blocks, and our analysis elucidates how $z_0$ changes through subsequent residual blocks.

\begin{figure*}[t]
    \centering
    \includegraphics[width=\linewidth]{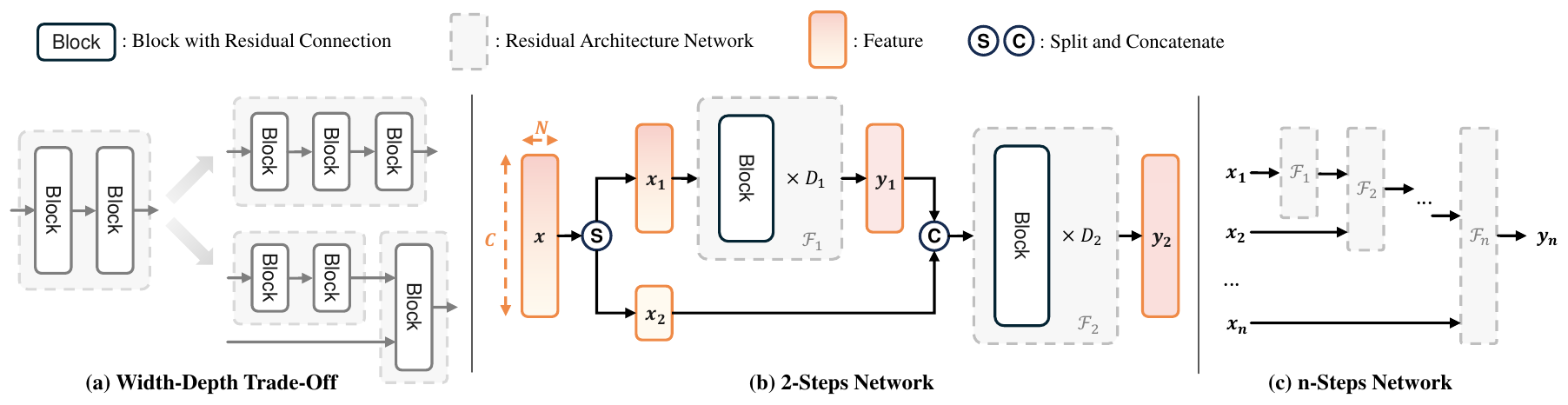}
    \vskip -0.2cm
    \caption{An illustration of the proposed Step by Step Network. For simplicity, the shortcut in each residual block is omitted.
    \textbf{(a) Width-depth trade-off.} When enlarging the depth of a residual model, the width has to be reduced to maintain similar computation. In contrast, StepsNet makes it possible for the model to be deeper with fixed width and computation.
    \textbf{(b) The 2-step network.} Given an input $x\in \mathbb{R}^{N\times C}$, the model first splits the information into two parts $x_1, x_2$ along the channel dimension $C$. Subsequently, $x_1$ and $x_2$ are processed sequentially by two networks $\mathcal{F}_1$ and $\mathcal{F}_2$, generating the final output $y_2$.  
    \textbf{(c) The $n$-step network.} Repeatedly substituting the first network with a 2-step architecture creates an $n$-step network, which divides $x$ into $n$ parts and processes them progressively.  }
    \label{fig:stepsnet}
    \vskip -0.3cm
\end{figure*}

\subsection{Limited Width}

Beyond optimization challenges posed by shortcut degradation, we further identify a fundamental architectural limitation arising from the depth-width trade-off under fixed computational budgets. Specifically, for a residual model $M$ with $D$ blocks of width $C$, we always have $\Omega(M)=\mathcal{O}(C^2D)$, where $\Omega(M)$ denotes the total computation complexity of $M$. For example, assume each block of model $M$ is a Transformer block (as shown in \cref{fig:transformer}, comprising an attention sub-block and an MLP sub-block), then we have:
\begin{equation} \label{eq:complexity}
    \begin{split}
        \Omega(M)\! =&(\underbrace{3NC^2\!\!+\!\!NC^2}_\text{\color{blue} QKV,\ Out\ Linear}\!+\underbrace{2N^2C}_\text{\color{blue} Attention}+\underbrace{8NC^2}_\text{\color{blue} MLP})\times D \\
        =&12NC^2D+2N^2CD =\mathcal{O}(C^2D),
    \end{split}
\end{equation}
where $N$ is the number of input tokens.

In deep learning, comparing models with similar computational budget $T$ is common practice, as bigger models (models with higher computation) always have better capacity. Given $\Omega(M)\!\!=\!\!\mathcal{O}(C^2D)$, with fixed budget $T$ and block design, the width $C$ will inevitably decrease proportionally as depth $D$ increases. 
For instance, \cref{fig:stepsnet}(a) illustrates a simple two-block residual network. To design a three-block model with the same computation, we have to narrow the model width by a factor of about $\sqrt{3/2}$.
Therefore, very deep residual models have a more limited width than their shallower counterparts with \textit{comparable computational budgets}.
As theoretical works~\cite{universal_approx, min_width, min_width_2} prove that networks with insufficient width can not achieve universal function approximation even with infinite depth, this limited dimensionality can be a key constraint in very deep models, hindering their capability when compared to shallower counterparts of similar computation.


\section{Step by Step Network}
\label{sec:method}



As discussed in \cref{sec:challenges}, despite their success, current residual architectures still face challenges at greater depths, potentially impeding the development of deeper networks.
Specifically, we present two theoretical observations:
\begin{itemize}
\vskip 0.05in
    \item \textit{The shortcut can ``degrade'' in very deep residual models.} The model adds too many noisy residual terms to the shortcut, preventing deep layers from accessing input information and propagating their gradients back during training. To mitigate this issue, early features need to be transmitted to deep layers more efficiently.

    \item \textit{The width-depth trade-off limits deep residual models to insufficient width.} More advanced design is needed to achieve large depth and sufficient width concurrently, without exceeding the computational budget.
    \vskip 0.05in
\end{itemize}

Motivated by these analyses, we introduce a progressive modeling strategy to alleviate both challenges simultaneously, presenting the \textbf{Step by Step Network (StepsNet)} as an improved generalization of the residual architecture.
For better understanding, we first present the StepsNet design in \cref{sec:stepsnet}, and then offer detailed analyses of how it addresses the two challenges in \cref{sec:how_address}.

\subsection{Let the Network think Step by Step}
\label{sec:stepsnet}

Given an input sequence $x\in \mathbb{R}^{N\times C}$, the architecture of proposed step by step network can be formulated as:
\begin{equation}
\label{eq:define_stepsnet}
    y_1=\mathcal{F}_1(x_1),\ \ y_2=\mathcal{F}_2([y_1, x_2]),
\end{equation}
where $x_1\!\in\!\mathbb{R}^{N\times d_1},x_2\!\in\!\mathbb{R}^{N\times d_2}$ are the two parts of $x$ split along the width dimension $C$, with $d_1\!+\!d_2\!=\!C$ and $x\!=\![x_1, x_2]$. $\mathcal{F}_1$ and $\mathcal{F}_2$ denote two residual architecture networks with widths $C_1\!=\!d_1$ and $C_2\!=\!d_1\!+\!d_2$, respectively. 

\cref{fig:stepsnet}(b) provides an illustration of this architecture. Unlike conventional residual architecture networks that process the entire input $x\in \mathbb{R}^{N\times C}$ directly, our method starts with \textit{a subspace of the input} $x_1\in\mathbb{R}^{N\times d_1}\subset \mathbb{R}^{N\times C}$, modeling the information in this subspace with $y_1=\mathcal{F}_1(x_1)\in\mathbb{R}^{N\times d_1}$. This intermediate result $y_1$ is then combined with $x_2$ to form the intermediate feature in \textit{the entire space}  $[y_1, x_2]\in \mathbb{R}^{N\times C}$, which is processed by $\mathcal{F}_2$ to obtain the final output $y_2=\mathcal{F}_2([y_1, x_2])\in\mathbb{R}^{N\times C}$.

Intuitively, the input $x$ contains diverse information. The proposed method explicitly directs the model to focus first on the first part of information $x_1$ and ``think one step'' with network $\mathcal{F}_1$, yielding some intermediate results $y_1$. On this basis, the second part of information $x_2$ is taken into consideration and the model can ``think another step'' to produce the final output based on $y_1$ and $x_2$. This step by step process shares some similarities with human problem-solving: when tackling complex problems, we often start with part of the conditions, derive preliminary insights, and then proceed to the remaining factors. Therefore, we refer to the proposed structure as Step by Step Network (StepsNet). Additionally, we term $x_1, x_2$ as \textit{slow path} and \textit{fast path}, respectively, since the information in $x_1$ takes two thinking steps while $x_2$ is directly considered by the second step.

Furthermore, the model defined by \cref{eq:define_stepsnet} can be viewed as a 2-step network, as there are two networks with different widths, $\mathcal{F}_1$ and $\mathcal{F}_2$. More generally, we define the $n$-step network as follows:
\begin{equation}
\label{eq:n_stepsnet}
    \begin{split}
        y_1&=\mathcal{F}_1(x_1), \\
        y_i&=\mathcal{F}_i([y_{i-1},x_i]),\ \ i=2,\cdots,n, \\
    \end{split}
\end{equation}
where $x_i\in\mathbb{R}^{N\times d_i}, \sum_{i=1}^{n}{d_i}=C,x=[x_1,\cdots,x_n],y_i\in\mathbb{R}^{N\times C_i}, C_i=\sum_{j=1}^{i}{d_j}$, and $y_n$ denotes the final output. Here, $\mathcal{F}_i$ represents the $i$-th step residual model with $D_i$ blocks of width $C_i$, $i=1,\cdots,n$. In practice, we set the width of each step to be $\sqrt{2}$ narrower than its subsequent step, \textit{i.e.} $C_i\!=\!C_{i+1}/\sqrt{2}$. The width of the whole network is $C$, matching the dimensions of $x,y_n$ and $\mathcal{F}_n$, and the total depth (in blocks) is $\sum_{i=1}^{n}D_i$. Interestingly, the proposed StepsNet practically forms a generalization of the residual architecture, where different blocks can have various widths, following a narrow-to-wide pattern. Furthermore, the conventional residual models can be viewed as 1-step variants of StepsNet.

\subsection{Going Deeper with StepsNet}
\label{sec:how_address}
We present a detailed analysis of how StepsNet overcomes the two challenges of very deep residual models:

\noindent
\textbf{The shortcut degradation.} As analyzed in \cref{sec:challenges}, very deep residual models introduce excessive noisy residuals into the shortcut, hindering optimization. StepsNet alleviates this problem with its progressive modeling strategy. 
Specifically, as shown in \cref{fig:stepsnet}(c), in the early steps of StepsNet, the residual terms are added only to a subspace of the entire input $x$. For example, $\mathcal{F}_1$ only processes the $x_1$ subspace, while $x_2, \cdots, x_n$ are kept uncorrupted. This effectively constrains the impact of residual terms and directly transmits the input information $x_2, \cdots, x_n$ to deep layers $\mathcal{F}_2, \cdots, \mathcal{F}_n$. 
We verify our analysis by calculating the shortcut ratio $\gamma_l$ in the deep Steps-DeiT model (482 layers). \cref{fig:std} shows that the value of $\gamma_l$ in the early layers is close to 1.0 and remains around 0.5 in deep layers, akin to the trend in the 122-layer DeiT. Therefore, StepsNet effectively overcomes shortcut degradation, allowing deep models to be optimized as easily as shallow residual networks.

\noindent
\textbf{The limited width.} 
As analyzed in \cref{sec:challenges}, residual architectures exhibit an inherent trade-off between width and depth. Under fixed computational constraints, residual models cannot be both wide and deep, resulting in limited width for extremely deep networks. In contrast, with our StepsNet design, it is possible to make a model deeper while fixing its width and computation. 
For example, consider the two-block residual network depicted in \cref{fig:stepsnet}(a). We could change the first block into two narrower blocks with equivalent computation while preserving the second block unchanged. In this way, the model is turned into a 2-step StepsNet with 3 blocks while maintaining the original width and computation. Moreover, applying this process recursively yields networks with $3, 4, \ldots, n$ steps. Therefore, the proposed StepsNet architecture allows for a significant model depth increase without sacrificing the entire model width or introducing computation overhead. This design effectively mitigates the width limitation of deep residual networks, enabling the construction of deep models with sufficient width within a fixed computational budget.

\section{Experiments}
\label{sec:exp}

Extremely deep networks are believed to have great expressive power but often fail to yield satisfactory results in practice. In \cref{sec:challenges} and \cref{sec:method}, we analyzed the challenges of current very deep networks, and proposed the StepsNet architecture as a possible solution for deep models. In this section, we conduct experimental studies to fully validate the effectiveness of our method.

\subsection{Setups}
The proposed StepsNet is a simple and clean macro architecture applicable to various models, regardless of their micro block designs. To assess its effectiveness, we implement StepsNet on three representative CNN and Transformer models, including ResNet~\cite{resnet}, DeiT\cite{deit}, and Swin Transformer~\cite{swin}. Comparisons with advanced methods are also included. We report experimental results on ImageNet-1K classification~\cite{imagenet}, COCO object detection~\cite{coco}, and ADE20K semantic segmentation~\cite{ade20k}. Additionally, we also apply StepsNet to language modeling tasks and conduct experiments using WikiText-103~\cite{wiki}. For a fair comparison, \textit{all experiment settings are consistent with baseline methods.} Please refer to the Appendix for details.

\begin{table}[t]
\centering
\footnotesize
\vskip 0.0cm
    \setlength{\tabcolsep}{1.0mm}{
    \renewcommand\arraystretch{1.05}
    \begin{tabular}{l|c c|c c|l}
        \toprule
        \textbf{Method} 
        & \textbf{\#Blocks} & \textbf{\#Layers} & \textbf{\#Params} & \textbf{FLOPs}    & \textbf{Top-1} \\

        \midrule
        ResNet-18~\cite{resnet}
        & 8                 & 18                & 11.7M             & 1.8G              & 70.2 \\
        ResNet-34
        & 16                & 34                & 21.8M             & 3.7G              & 75.0 \\
        ResNet-50
        & 16                & 50                & 25.6M             & 4.1G              & 78.8 \\

        \midrule
        \rowcolor{lightgray!20} Steps-ResNet-18
        & 18                & 38                & 11.7M             & 1.8G              & 71.8 \\
        \rowcolor{lightgray!20} Steps-ResNet-34
        & 34                & 70                & 21.8M             & 3.7G              & 76.0 \\
        \rowcolor{lightgray!20} Steps-ResNet-50
        & 34                & 104               & 25.6M             & 4.1G              & 79.9 \\
        
        \midrule
        DeiT-T~\cite{deit}
        & 12                & 62                & 5.7M              & 1.3G              & 72.2 \\
        DeiT-S
        & 12                & 62                & 22.1M             & 4.6G              & 79.9 \\
        DeiT-B
        & 12                & 62                & 86.6M             & 17.6G             & 81.8 \\

        \midrule
        \rowcolor{lightgray!20} Steps-DeiT-T
        & 24                & 122               & 5.7M              & 1.3G              & 73.2 \\
        \rowcolor{lightgray!20} Steps-DeiT-S
        & 24                & 122               & 22.1M             & 4.7G              & 81.0 \\
        \rowcolor{lightgray!20} Steps-DeiT-B
        & 24                & 122               & 86.7M             & 17.9G             & 82.7 \\

        \midrule
        Swin-T~\cite{swin}
        & 12                & 65                & 28.3M             & 4.5G              & 81.3 \\
        Swin-S
        & 24                & 125               & 49.6M             & 8.8G              & 83.0 \\
        Swin-B
        & 24                & 125               & 87.8M             & 15.5G             & 83.5 \\

        \midrule
        \rowcolor{lightgray!20} Steps-Swin-T
        & 26                & 135               & 27.8M             & 4.5G              & 82.4 \\
        \rowcolor{lightgray!20} Steps-Swin-S
        & 38                & 195               & 49.1M             & 8.8G              & 83.6 \\
        \rowcolor{lightgray!20} Steps-Swin-B
        & 38                & 195               & 86.7M             & 15.5G             & 84.1 \\

        \bottomrule
    \end{tabular}}
\vskip -0.2cm
\caption{Comparison of different models on ImageNet-1K. Consistent with ResNet~\cite{resnet}, we count each convolutional or linear layer as one layer. Therefore, each Transformer block (see \cref{fig:transformer}) is counted as $3(\mathrm{QKV, Attention, Out}) + 2(\mathrm{MLP}) = 5$ layers.}
\vskip -0.3cm
\label{tab:cls}
\end{table}

\subsection{Main Results and Broad Comparisons}
\label{sec:main_result}

In this section, we conduct comprehensive experiments to fully evaluate the effectiveness of StepsNet across various model sizes and tasks. 

\noindent
\textbf{ImageNet classification.}
We replace the residual architecture of ResNet~\cite{resnet}, DeiT~\cite{deit}, and Swin Transformer~\cite{swin} with our StepsNet, without any other modifications. \textit{Please see detailed model architectures in the Appendix. }

The results are presented in \cref{tab:cls}. StepsNet achieves consistent improvements against baseline models. For instance, our Steps-Swin-T surpasses Swin-T by 1.1\% in accuracy, and Steps-Swin-S outperforms Swin-B while using only 56\% of the parameters and FLOPs. The 302-layer Steps-DeiT-B model matches the performance of DeiT-B with 50\% fewer parameters and FLOPs. These results suggest that StepsNet tends to be a superior alternative to the widely adopted residual architecture. Furthermore, the results also demonstrate the incredible power of network depth, as our method introduces no additional operations but solely benefits the model with increased depth.

\begin{figure}[t]
    \centering
    \includegraphics[width=0.8\linewidth]{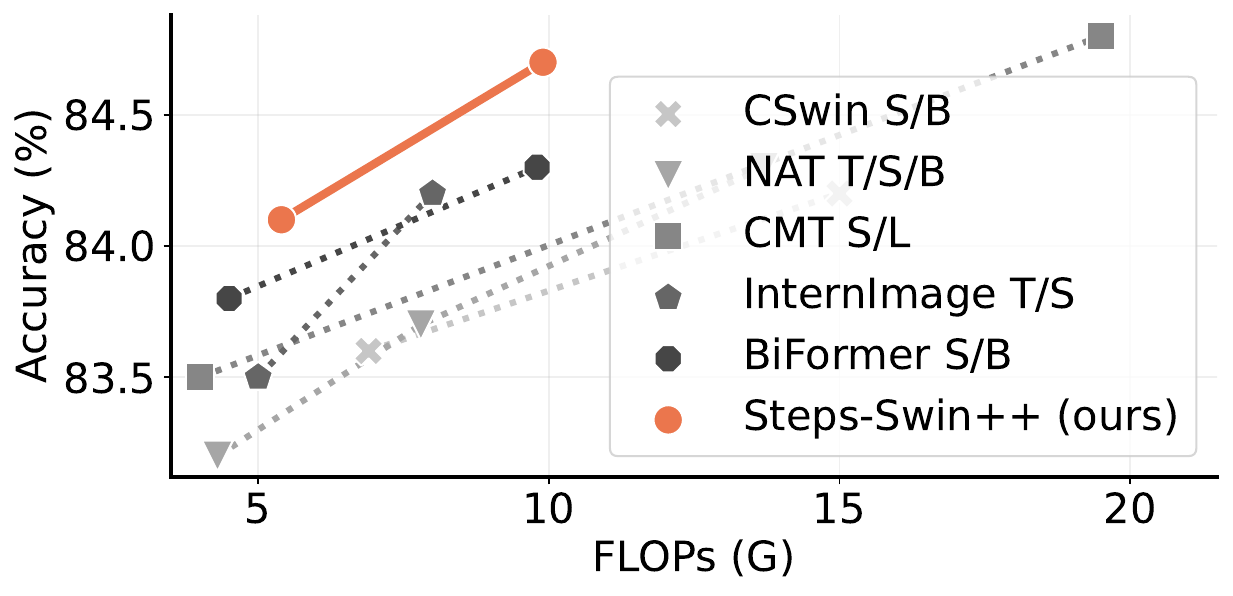}
    \vskip -0.2cm
    \caption{Comparison with advanced methods on ImageNet-1K. }
    \vskip -0.2cm
    \label{fig:cls_sota}
\end{figure}

To further explore the capacity of StepsNet and compare with advanced methods, we integrate components commonly used in advanced models~\cite{cmt, biformer} like LPU~\cite{cmt} and ConvFFN~\cite{cmt} into the baseline Swin model. As shown in \cref{fig:cls_sota}, the enhanced Steps-Swin++ models achieve highly competitive results and surpass various advanced designs, highlighting the superior capacity of our method.

\begin{table}[t]
\centering
\footnotesize
\setlength{\tabcolsep}{0.7mm}{
\renewcommand\arraystretch{1.05}
\begin{tabular}{l|c|c|ccc|ccc}
    \toprule
    \multicolumn{9}{c}{\textbf{(a) Mask R-CNN Object Detection on COCO}} \\
    Method & FLOPs & Sch. & AP$^b$ & AP$^b_\text{50}$ & AP$^b_\text{75}$ & AP$^m$ & AP$^m_\text{50}$ & AP$^m_\text{75}$ \\
    
    \hline Swin-T~\cite{swin}
    & 267G & 1x      & 43.7 & 66.6 & 47.7 & 39.8 & 63.3 & 42.7 \\
    \rowcolor{lightgray!20} Steps-Swin-T
    & 266G & 1x      & 44.1 & 67.0 & 48.3 & 40.2 & 63.8 & 42.9 \\

    \hline Swin-S
    & 359G & 1x      & 45.7 & 67.9 & 50.4 & 41.1 & 64.9 & 44.2 \\
    \rowcolor{lightgray!20} Steps-Swin-S
    & 357G & 1x      & 46.3 & 68.8 & 50.9 & 41.9 & 65.7 & 45.0 \\

    \hline Swin-B
    & 503G & 1x      & 46.9 & -    & -    & 42.3 & -    & -    \\
    \rowcolor{lightgray!20} Steps-Swin-B
    & 501G & 1x      & 47.1 & 69.6 & 51.6 & 42.5 & 66.6 & 45.6 \\

    \toprule
    
    \multicolumn{9}{c}{\textbf{(b) Cascade Mask R-CNN Object Detection on COCO}} \\
    Method & FLOPs & Sch. & AP$^b$ & AP$^b_\text{50}$ & AP$^b_\text{75}$ & AP$^m$ & AP$^m_\text{50}$ & AP$^m_\text{75}$ \\
    
    \hline Swin-T~\cite{swin}
    & 745G & 3x      & 50.4 & 69.2 & 54.7 & 43.7 & 66.6 & 47.3 \\
    \rowcolor{lightgray!20} Steps-Swin-T
    & 745G & 3x      & 51.2 & 70.1 & 55.8 & 44.3 & 67.4 & 48.3 \\
    
    \hline Swin-S 
    & 837G & 3x      & 51.9 & 70.7 & 56.3 & 45.0 & 68.2 & 48.8 \\
    \rowcolor{lightgray!20} Steps-Swin-S
    & 836G & 3x      & 52.3 & 71.1 & 57.0 & 45.2 & 68.5 & 49.1 \\

    \hline Swin-B 
    & 981G & 3x      & 51.9 & 70.5 & 56.4 & 45.0 & 68.1 & 48.9 \\
    \rowcolor{lightgray!20} Steps-Swin-B
    & 979G & 3x      & 52.5 & 71.4 & 57.2 & 45.4 & 69.1 & 49.0 \\
    
    \toprule
\end{tabular}}
\vskip -0.2cm
\caption{Results on COCO dataset. The FLOPs are computed over the backbone, FPN, and detection head with a $1280\times800$ input.}
\label{tab:det}
\vskip -0.2cm
\end{table}

\begin{table}[t]
\centering
\footnotesize
\setlength{\tabcolsep}{1.5mm}{
\renewcommand\arraystretch{1.2}
\begin{tabular}{l|c|c c|c c}
    \toprule
    \multicolumn{6}{c}{\textbf{Semantic Segmentation on ADE20K}} \\
    Backbone & Method & FLOPs & \#Params & mIoU & mAcc \\
    
    \hline Swin-T~\cite{swin}
    & UperNet & 945G & 60M & 44.5 & 55.6 \\
    \rowcolor{lightgray!20} Steps-Swin-T
    & UperNet & 941G & 58M & 45.5 & 56.8 \\
    \toprule
\end{tabular}}
\vskip -0.2cm
\caption{Results of semantic segmentation. The FLOPs are computed over encoders and decoders with a 512$\times$2048 input.}
\vskip -0.2cm
\label{tab:seg}
\end{table}

\begin{table}[t]
\centering
\footnotesize
\setlength{\tabcolsep}{1.5mm}{
\renewcommand\arraystretch{1.05}
\begin{tabular}{l|c|c c|c}
    \toprule
    \multicolumn{5}{c}{\textbf{Language Modeling on WikiText-103}} \\
    Backbone & \#Params & \#Blocks & \#Layers & PPL$\downarrow$ \\
    \hline 
    Transformer~\cite{llama}
    & 30M & 6 & 31 & 25.28 \\
    \rowcolor{lightgray!20} Steps-Transformer
    & 30M & 12 & 61 & 24.39 \\
    \hline
    Transformer 
    & 44M & 6 & 31 & 24.45 \\
    \rowcolor{lightgray!20} Steps-Transformer
    & 45M & 12 & 61 & 24.00 \\
    \hline
    Transformer
    & 63M & 12 & 61 & 22.05 \\
    \rowcolor{lightgray!20} Steps-Transformer
    & 64M & 24 & 121 & 21.64 \\
    \toprule
\end{tabular}}
\vskip -0.2cm
\caption{Results of language modeling perplexity on WikiText-103. Parameters from the word embedding layer are included in the total parameter count.}
\vskip -0.3cm
\label{tab:nlp}
\end{table}

\noindent
\textbf{COCO object detection.}
In \cref{tab:det}, we provide the object detection results of different model sizes and detection heads. Consistent with the trend observed in classification, StepsNet enables the model to benefit from increased depth, delivering improved results in all settings. This validates its effectiveness in dense prediction scenarios. 

\noindent
\textbf{ADE20K semantic segmentation.}
We report the results on ADE20K dataset in Tab. 5. Similar to the object detection task, our method yields better results in semantic segmentation, further verifying the effectiveness of StepsNet architecture and the importance of depth.

\noindent
\textbf{Language modeling.}
Recent research~\cite{physics} suggests that deeper language models can achieve superior performance in reasoning tasks compared to their shallower counterparts, calling for the exploration of model depth in language modeling.
As shown in \cref{tab:nlp}, we apply StepsNet architecture to decoder-only Transformer~\cite{llama} to empirically assess its effectiveness in language modeling. The models are trained on WikiText-103.
In terms of language modeling perplexity, StepsNet consistently outperform vanilla Transformer across different model sizes. This confirms the effectiveness of our design in language modeling task and demonstrates the benefits of increased model depth.

\begin{figure}[t]
    \centering
    \includegraphics[width=\linewidth]{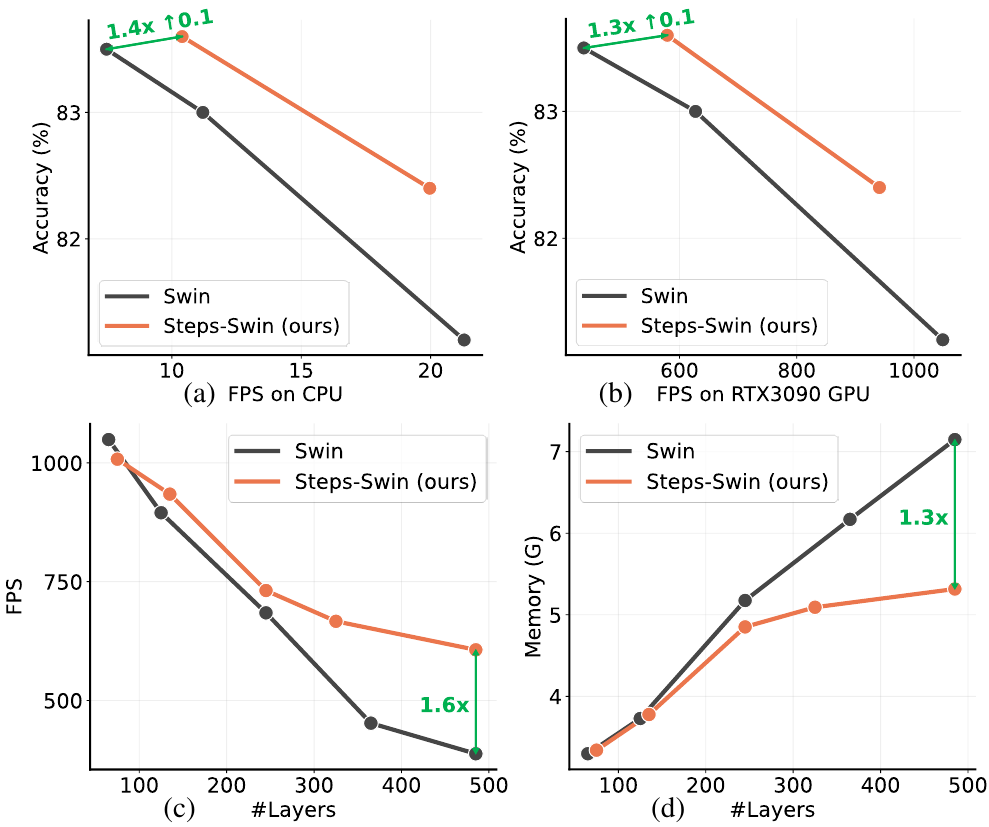}
    \vskip -0.2cm
    \caption{Speed and memory measurements. (a, b) Throughput-accuracy curves on CPU and RTX3090 GPU. (c) Throughput for models of varying depths, tested on an RTX3090 GPU. (d) Memory usage of models with different depths is measured as the change in total memory when batch size increases from 32 to 64. }
    \vskip -0.3cm
    \label{fig:speed}
\end{figure}

\noindent
\textbf{Throughput and memory.}
We further present real speed and memory usage measurements in \cref{fig:speed}. StepsNet achieves a better trade-off between throughput and accuracy, providing a 1.3x-1.4x speedup with improved performance on both CPU and RTX3090 GPU. In \cref{fig:speed}(c) and (d), we illustrate the throughput and memory usage of models in \cref{tab:deeper}(b). In residual architectures, deeper models tend to exhibit lower throughput and increased memory usage. Our design mitigates this issue, enabling a 1.6x speedup and 1.3x less memory usage at large depths.

\subsection{Building Deeper Models}
\label{sec:exp_deeper}

In \cref{sec:main_result}, we validated our method using existing residual models. In this section, we develop deeper models at the DeiT-T/S scale ($1.0\!\sim\!5.0$G FLOPs) to further evaluate StepsNet at greater depths with manageable computational costs. We conduct two series of experiments:
\begin{itemize}
\vskip 0.05in
    \item \textit{Deeper models with fixed width.} The standard DeiT-T model employs $192$ width and $12$ blocks. To build deeper model at DeiT-T scale, we first reduce the width to $136$ (by a factor of $\sqrt{2}$) while doubling the depth to $24$ blocks. Based on this model, we progressively increase model depth to 96 blocks while keeping width fixed, establishing the baseline results. To ensure a fair comparison, we set the width of Steps-DeiT to $136\times 2^{\frac{i-5}{2}}, i=1,\cdots,7$, maintaining a computational cost similar to the corresponding baseline model at the same depth.
\vskip 0.05in
    \item \textit{Deeper models with fixed FLOPs.} We start from the standard Swin-T and hold FLOPs constant, incrementally increasing model depth while reducing the width. This allows a controlled examination of width-depth trade-off. The Steps-Swin-T is configured similarly.
\vskip 0.05in
\end{itemize}

\noindent
\textbf{The results}
are provided in \cref{tab:deeper} and \cref{fig:deeper}. 
In \textit{the first setup}, where width is fixed, the residual architecture DeiT exhibits modest accuracy gains beyond 200 layers. Notably, this is not mainly due to overfitting, as similar trends also appear in training accuracy (see \cref{fig:deeper}). We attribute this to the optimization difficulties brought by the shortcut degradation analyzed in \cref{sec:challenges}. In contrast, our design alleviates this problem, enabling Steps-DeiT to achieve increasingly better results at greater depth. 
In \textit{the second setup}, we can observe a clear width-depth trade-off under fixed FLOPs in residual architectures. Specifically, Swin model achieves peak performance around 120 layers but degrades significantly as depth further increases. As discussed in \cref{sec:method}, StepsNet mitigates this trade-off, helping the model benefit from depth more effectively. This is verified by the results that Steps-Swin model consistently outperforms across all depths and achieves fairly strong accuracy at great depth.
These results support our analyses in \cref{sec:challenges} and \cref{sec:method}, and demonstrate the effectiveness of StepsNet.

\begin{table}[t]
\centering
\footnotesize
\setlength{\tabcolsep}{0.9mm}{
\renewcommand\arraystretch{1.05}
\begin{tabular}{c c|c c|c c c c|c c|c}
    \toprule
    \multicolumn{11}{c}{\textbf{(a) Going Deeper with Fixed Width}} \\
    \multicolumn{5}{c}{DeiT~\cite{deit}} & & \multicolumn{5}{c}{Steps-DeiT} \\
    \cline{1-5} \cline{7-11}
    \#B & \#L & \#Params & FLOPs & Acc. &  & \#B & \#L & \#Params & FLOPs & Acc. \\
    
    \cline{1-5} \cline{7-11}
    24 & 122 & 5.6M & 1.3G & 72.9 & & 24 & 122 & 5.9M & 1.3G & 73.3 \\
    36 & 182 & 8.3M & 2.0G & 75.2 & & 36 & 182 & 8.8M & 2.0G & 77.0 \\
    48 & 242 & 11.0M & 2.6G & 75.7 & & 48 & 242 & 11.3M & 2.6G & 78.4 \\
    72 & 362 & 16.4M & 4.0G & 75.8 & & 72 & 362 & 16.6M & 3.8G & 79.4 \\
    96 & 482 & 21.8M & 5.3G & 75.6 & & 96 & 482 & 22.0M & 5.1G & 80.0 \\
    
    \toprule
    
    \multicolumn{11}{c}{\textbf{(b) Going Deeper with Fixed FLOPs}} \\
    \multicolumn{5}{c}{Swin~\cite{swin}} & & \multicolumn{5}{c}{Steps-Swin} \\
    \cline{1-5} \cline{7-11}
    \#B & \#L & \#Params & FLOPs & Acc. & & \#B & \#L & \#Params & FLOPs & Acc. \\
    
    \cline{1-5} \cline{7-11}
    12 & 65 & 28.3M & 4.5G & 81.3 & & 14 &75 & 27.8M & 4.5G & 81.8 \\
    24 & 125 & 25.1M & 4.5G & 81.8 & & 26 & 135 & 27.8M & 4.5G & 82.4 \\
    48 & 245 & 24.5M & 4.5G & 80.9 & & 48 & 245 & 28.8M & 4.6G & 82.4 \\
    72 & 365 & 24.1M & 4.5G & 80.0 & & 64 & 325 & 28.8M & 4.6G & 82.2 \\
    96 & 485 & 24.3M & 4.6G & 79.8 & & 96 & 485 & 28.8M & 4.6G & 82.1 \\
    
    \toprule
\end{tabular}}
\vskip -0.2cm
\caption{Deeper model test accuracy on ImageNet. \#B and \#L denote number of blocks and layers, respectively.}
\label{tab:deeper}
\vskip -0.3cm
\end{table}

\subsection{Analysis of StepsNet}
In this section, we offer additional analysis and discussion of StepsNet's behavior and effectiveness. 

\noindent
\textbf{Slow and fast path.} 
We offer analytical results to demystify the role of slow and fast paths (defined in \cref{sec:stepsnet}) in StepsNet. Specifically, we employ the pretrained 3-step Steps-DeiT-S, setting $n$ channels of the slow ($x_1$) or fast path ($x_2, x_3$) to zero, and directly evaluating the model’s accuracy. The results are depicted in \cref{tab:mask_out_channels}. Setting channels in the slow path to zero leads to a significant performance drop, whereas masking fast path channels has a relatively minor effect. This indicates that StepsNet can learn to route essential information to the slow path to process it with more blocks and steps, while placing complementary information in fast channels to consider them later.


\begin{table}[t]
    \centering
    \vskip -0.2cm
    \footnotesize
    \setlength{\tabcolsep}{1.0mm}{
    \renewcommand\arraystretch{1.05}
    \begin{tabular}{c|c c c c c c}
        Mask Out Channels   & None & 32 & 64 & 96 & 128 & 160 \\
        \midrule
        Slow Path           & 81.0 & 77.3 & 68.3 & 49.0 & 20.4 & 4.4   \\
        Fast Path           & 81.0 & 80.9 & 80.9 & 80.8 & 80.7 & 80.3   \\
    \end{tabular}}
    \vskip -0.2cm
    \caption{Performances on ImageNet-1K when masking out the first or last $n$ channels of input $x$ before the StepsNet architecture in Steps-DeiT-S. Models are tested without retraining. }
    \label{tab:mask_out_channels}
    \vskip -0.2cm
\end{table}

\noindent
\textbf{Contribution of each step.} 
We drop similar computations in $\mathcal{F}_1,\mathcal{F}_2,\mathcal{F}_3$ of the pretrained Steps-DeiT-S to assess the contribution of similar computations in different steps. Specifically, we drop the last 4 blocks in $\mathcal{F}_1$, 2 blocks in $\mathcal{F}_2$, or 1 block in $\mathcal{F}_3$ (as they have similar computations), and evaluate the model’s accuracy. \cref{tab:drop_blocks} shows that dropping similar computations in earlier steps leads to a more significant performance drop. This suggests that similar computations play a more important role when placed in narrower networks like $\mathcal{F}_1$. Therefore, changing the 1-step residual model into multi-step StepsNet enables similar computations to perform better, leading to StepsNet's effectiveness.

\begin{table}[t]
    \centering
    \vskip -0.2cm
    \footnotesize
    \setlength{\tabcolsep}{1.5mm}{
    \renewcommand\arraystretch{1.05}
    \begin{tabular}{c|c c c}
        Dropping Position  & $\mathcal{F}_1$ & $\mathcal{F}_2$ & $\mathcal{F}_3$ \\
        \midrule
        Acc.               & 77.0 & 78.1 & 80.5   \\
    \end{tabular}}
    \vskip -0.2cm
    \caption{Dropping similar computations in the three networks of Steps-DeiT-S. Models are tested without retraining. }
    \label{tab:drop_blocks}
    \vskip -0.2cm
\end{table}

\subsection{Ablation Studies}

In this section, we conduct ablation studies of the key designs in StepsNet. We report ImageNet-1K classification results using Steps-DeiT-S and Steps-Swin-T.

\noindent
\textbf{The number of steps.} 
In \cref{tab:ablation_steps}, we examine the effect of increasing the number of steps in StepsNet. Results indicate that even a 2-step network outperforms the baseline residual model (equivalent to 1-step StepsNet). Appropriately introducing more steps can further benefit the model with depth. In practice, we choose 3 steps as the default, which doubles the model depth with little computation overhead.

\noindent
\textbf{Computation allocation.} 
We further evaluate the computation allocation of StepsNet in \cref{tab:ablation_allocation}. Steps-DeiT-S comprises three networks, $\mathcal{F}_1,\mathcal{F}_2,\mathcal{F}_3$, with widths $\frac{C}{2},\frac{C}{\sqrt{2}},C$ and depths $D_1,D_2,D_3$, respectively. We gradually reduce the depth of the widest sub-model $\mathcal{F}_3$ and correspondingly increase the depths of $\mathcal{F}_1,\mathcal{F}_2$ to maintain similar complexity. We find that a simple $(12, 6, 6)$ allocation is effective, which is achieved by allocating half of each subsequent network’s computation to the previous one.

\begin{table}[t]
    \centering
    \footnotesize
    \setlength{\tabcolsep}{1.6mm}{
    \renewcommand\arraystretch{1.05}
    \begin{tabular}{c|c c|c c|c}
        \#Steps & \#Blocks  & \#Layers  & \#Param   & FLOPS     & Acc. \\
        \midrule
        1       & 12        & 62        & 22.1M     & 4.6G      & 79.9  \\
        2       & 18        & 92        & 22.1M     & 4.7G      & 80.8  \\
        \rowcolor{lightgray!20}
        3       & 24        & 122       & 22.1M     & 4.7G      & 81.0  \\
        4       & 30        & 152       & 22.1M     & 4.8G      & 81.0  \\
        5       & 36        & 182       & 22.1M     & 4.8G      & 81.2  \\
    \end{tabular}}
    \vskip -0.2cm
    \caption{Ablation on the number of steps.}
    \label{tab:ablation_steps}
    \vskip -0.2cm
\end{table}

\begin{table}[t]
    \centering
    \footnotesize
    \setlength{\tabcolsep}{1.5mm}{
    \renewcommand\arraystretch{1.05}
    \begin{tabular}{c|c c|c c|c}
        $(D_1, D_2, D_3)$   & \#Blocks  & \#Layers  & \#Param   & FLOPS     & Acc. \\
        \midrule
        $(0, 0, 12)$        & 12        & 62        & 22.1M     & 4.6G      & 79.9 \\
        $(2, 1, 11)$        & 14        & 72        & 22.1M     & 4.6G      & 80.5  \\
        $(6, 3, 9)$         & 18        & 92        & 22.1M     & 4.7G      & 80.8  \\
        \rowcolor{lightgray!20}
        $(12, 6, 6)$        & 24        & 122       & 22.1M     & 4.7G      & 81.0  \\
        $(18, 9, 3)$        & 30        & 152       & 22.1M     & 4.8G      & 81.0  \\
        $(22, 11, 1)$       & 34        & 172       & 22.1M     & 4.9G      & 81.0  \\
    \end{tabular}}
    \vskip -0.2cm
    \caption{Ablation on computation allocation.}
    \label{tab:ablation_allocation}
    \vskip -0.2cm
\end{table}

\noindent
\textbf{The reverse design.} 
StepsNet uses a narrow-to-wide stacking strategy, where the width of each network $\mathcal{F}_i$ increases with $i$, as shown in \cref{fig:stepsnet}(c).
One reverse design is stacking the blocks wide to narrow, which can be formulated as:
\begin{equation}
\label{eq:reverse_stepsnet}
    \begin{split}
        x_{i+1},y_i&=\mathrm{Split}(\mathcal{F}_i(x_i)),\ \ i=1,\cdots,n-1, \\
        y_n&=\mathcal{F}_n(x_n), \\
    \end{split}
\end{equation}
where $x_1\in\mathbb{R}^{N\times C}$ denotes the input and $[y_1,\cdots,y_n]\in\mathbb{R}^{N\times C}$ is the output. The results in \cref{tab:ablation_key_designs} show that StepsNet significantly outperforms the reverse design despite having identical parameters and FLOPs. We attribute this disparity to different model depths. 
In StepsNet, although $x_i$ passes only through $\mathcal{F}_j,j=i,\cdots,n$, it can integrate with the features from all previous layers $y_{i-1}$ through $y_i=\mathcal{F}_i([y_{i-1},x_i])$ (see \cref{eq:n_stepsnet}). Therefore, each output channel of StepsNet contains the deepest feature from $x_1$, processed by all sub-models $\mathcal{F}_i,i=1,\cdots,n$. In the reverse design, however, only $y_n$ passes through all layers, while $y_i,i=1,\cdots,n-1$ are comparatively shallow. Therefore, the results validate the effectiveness of our narrow-to-wide stacking strategy and confirm that StepsNet benefits the model with increased depth. 

\begin{table}[t]
    \centering
    \footnotesize
    \setlength{\tabcolsep}{1.0mm}{
    \renewcommand\arraystretch{1.05}
    \begin{tabular}{c|c c|c c|c}
        Name            & \#Blocks  & \#Layers  & \#Param   & FLOPS     & Acc. \\
        \midrule
        Reverse         & 26        & 135       & 27.8M     & 4.5G      & 81.3  \\
        \rowcolor{lightgray!20}
        Steps-Swin-T    & 26        & 135       & 27.8M     & 4.5G      & 82.4 \\
    \end{tabular}}
    \vskip -0.2cm
    \caption{Ablation on the key designs of StepsNet.}
    \label{tab:ablation_key_designs}
    \vskip -0.3cm
\end{table}

\section{Conclusion}

Developing deeper, more expressive networks is a fundamental research topic in deep learning. Despite great advances, modern residual networks still struggle to deliver anticipated results at large depths. In this paper, we identify two key challenges of deep residual models: shortcut degradation and limited width. Based on these analyses, we introduce a generalized and improved residual architecture, Step by Step Network (StepsNet), as a possible solution for deep models. Our method effectively alleviates the identified limitations and is compatible with various models and micro designs. Empirical validations across various tasks confirm that StepsNet benefits models with increased depth, suggesting it as a simple, superior, and scalable generalization of the widely adopted residual architectures.

\section*{Acknowledgement}
This work is supported in part by the National Key R\&D Program of China under Grant 2024YFB4708200, the National Natural Science Foundation of China under Grants U24B20173 and 62276150, and the Scientific Research Innovation Capability Support Project for Young Faculty under Grant ZYGXQNJSKYCXNLZCXM-I20.

{
    \small
    \bibliographystyle{ieeenat_fullname}
    \bibliography{main}

@String(IJCV = {Int. J. Comput. Vis.})

@String(CVPR= {IEEE Conf. Comput. Vis. Pattern Recog.})

@String(ICCV= {Int. Conf. Comput. Vis.})

@String(ECCV= {Eur. Conf. Comput. Vis.})

@String(ICLR = {Int. Conf. Learn. Represent.})

@String(AAAI = {AAAI})

@String(CVPRW= {IEEE Conf. Comput. Vis. Pattern Recog. Worksh.})

@String(IJCV  = {IJCV})

@String(CVPR  = {CVPR})

@String(ICCV  = {ICCV})

@String(ECCV  = {ECCV})

@String(ICLR  = {ICLR})

@String(CVPRW= {CVPRW})

@inproceedings{deit,
  title={Training data-efficient image transformers \& distillation through attention},
  author={Touvron, Hugo and Cord, Matthieu and Douze, Matthijs and Massa, Francisco and Sablayrolles, Alexandre and J{\'e}gou, Herv{\'e}},
  booktitle={ICML},
  year={2021}
}

@inproceedings{pvt,
  title={Pyramid vision transformer: A versatile backbone for dense prediction without convolutions},
  author={Wang, Wenhai and Xie, Enze and Li, Xiang and Fan, Deng-Ping and Song, Kaitao and Liang, Ding and Lu, Tong and Luo, Ping and Shao, Ling},
  booktitle={ICCV},
  year={2021}
}

@inproceedings{swin,
  title={Swin transformer: Hierarchical vision transformer using shifted windows},
  author={Liu, Ze and Lin, Yutong and Cao, Yue and Hu, Han and Wei, Yixuan and Zhang, Zheng and Lin, Stephen and Guo, Baining},
  booktitle={ICCV},
  year={2021}
}

@inproceedings{vit,
  title={An image is worth 16x16 words: Transformers for image recognition at scale},
  author={Dosovitskiy, Alexey and Beyer, Lucas and Kolesnikov, Alexander and Weissenborn, Dirk and Zhai, Xiaohua and Unterthiner, Thomas and Dehghani, Mostafa and Minderer, Matthias and Heigold, Georg and Gelly, Sylvain and others},
  booktitle={ICLR},
  year={2021}
}

@inproceedings{maskformer,
  title={Per-pixel classification is not all you need for semantic segmentation},
  author={Cheng, Bowen and Schwing, Alex and Kirillov, Alexander},
  booktitle={NeurIPS},
  year={2021}
}

@inproceedings{detr,
  title={End-to-end object detection with transformers},
  author={Carion, Nicolas and Massa, Francisco and Synnaeve, Gabriel and Usunier, Nicolas and Kirillov, Alexander and Zagoruyko, Sergey},
  booktitle={ECCV},
  year={2020}
}

@inproceedings{segformer,
  title={SegFormer: Simple and efficient design for semantic segmentation with transformers},
  author={Xie, Enze and Wang, Wenhai and Yu, Zhiding and Anandkumar, Anima and Alvarez, Jose M and Luo, Ping},
  booktitle={NeurIPS},
  year={2021}
}

@inproceedings{vitdet,
  title={Exploring plain vision transformer backbones for object detection},
  author={Li, Yanghao and Mao, Hanzi and Girshick, Ross and He, Kaiming},
  booktitle={ECCV},
  year={2022}
}

@inproceedings{cmt,
  title={Cmt: Convolutional neural networks meet vision transformers},
  author={Guo, Jianyuan and Han, Kai and Wu, Han and Tang, Yehui and Chen, Xinghao and Wang, Yunhe and Xu, Chang},
  booktitle={CVPR},
  year={2022}
}

@inproceedings{mobilenetv2,
  title={Mobilenetv2: Inverted residuals and linear bottlenecks},
  author={Sandler, Mark and Howard, Andrew and Zhu, Menglong and Zhmoginov, Andrey and Chen, Liang-Chieh},
  booktitle={CVPR},
  year={2018}
}

@inproceedings{imagenet,
  title={Imagenet: A large-scale hierarchical image database},
  author={Deng, Jia and Dong, Wei and Socher, Richard and Li, Li-Jia and Li, Kai and Fei-Fei, Li},
  booktitle={CVPR},
  year={2009}
}

@inproceedings{adamw,
  title={Decoupled weight decay regularization},
  author={Loshchilov, Ilya and Hutter, Frank},
  booktitle={ICLR},
  year={2018}
}

@inproceedings{randaugment,
  title={Randaugment: Practical automated data augmentation with a reduced search space},
  author={Cubuk, Ekin D and Zoph, Barret and Shlens, Jonathon and Le, Quoc V},
  booktitle={CVPRW},
  year={2020}
}

@inproceedings{cutmix,
  title={Cutmix: Regularization strategy to train strong classifiers with localizable features},
  author={Yun, Sangdoo and Han, Dongyoon and Oh, Seong Joon and Chun, Sanghyuk and Choe, Junsuk and Yoo, Youngjoon},
  booktitle={ICCV},
  year={2019}
}

@inproceedings{mixup,
    title={mixup: Beyond Empirical Risk Minimization},
    author={Hongyi Zhang and Moustapha Cisse and Yann N. Dauphin and David Lopez-Paz},
    booktitle={ICLR},
    year={2018}
}

@inproceedings{random_erasing,
  title={Random erasing data augmentation},
  author={Zhong, Zhun and Zheng, Liang and Kang, Guoliang and Li, Shaozi and Yang, Yi},
  booktitle={AAAI},
  year={2020}
}

@inproceedings{coco,
  title={Microsoft coco: Common objects in context},
  author={Lin, Tsung-Yi and Maire, Michael and Belongie, Serge and Hays, James and Perona, Pietro and Ramanan, Deva and Doll{\'a}r, Piotr and Zitnick, C Lawrence},
  booktitle={ECCV},
  year={2014},
}

@article{ade20k,
  title={Semantic understanding of scenes through the ade20k dataset},
  author={Zhou, Bolei and Zhao, Hang and Puig, Xavier and Xiao, Tete and Fidler, Sanja and Barriuso, Adela and Torralba, Antonio},
  journal={IJCV},
  year={2019}
}

@inproceedings{mrcn,
  title={Mask r-cnn},
  author={He, Kaiming and Gkioxari, Georgia and Doll{\'a}r, Piotr and Girshick, Ross},
  booktitle={ICCV},
  year={2017}
}

@inproceedings{crcnn,
  title={Cascade r-cnn: Delving into high quality object detection},
  author={Cai, Zhaowei and Vasconcelos, Nuno},
  booktitle={CVPR},
  year={2018}
}

@inproceedings{upernet,
  title={Unified perceptual parsing for scene understanding},
  author={Xiao, Tete and Liu, Yingcheng and Zhou, Bolei and Jiang, Yuning and Sun, Jian},
  booktitle={ECCV},
  year={2018}
}

@inproceedings{attention,
  title={Attention is all you need},
  author={Vaswani, Ashish and Shazeer, Noam and Parmar, Niki and Uszkoreit, Jakob and Jones, Llion and Gomez, Aidan N and Kaiser, {\L}ukasz and Polosukhin, Illia},
  booktitle={NeurIPS},
  year={2017}
}

@InProceedings{flatten,
  title={FLatten Transformer: Vision Transformer using Focused Linear Attention},
  author={Han, Dongchen and Pan, Xuran and Han, Yizeng and Song, Shiji and Huang, Gao},
  booktitle={ICCV},
  year={2023}
}

@inproceedings{clip,
  title = 	 {Learning Transferable Visual Models From Natural Language Supervision},
  author =       {Radford, Alec and Kim, Jong Wook and Hallacy, Chris and Ramesh, Aditya and Goh, Gabriel and Agarwal, Sandhini and Sastry, Girish and Askell, Amanda and Mishkin, Pamela and Clark, Jack and Krueger, Gretchen and Sutskever, Ilya},
  booktitle = 	 {ICML},
  year = 	 {2021}
}

@inproceedings{biformer,
  title={BiFormer: Vision Transformer with Bi-Level Routing Attention},
  author={Zhu, Lei and Wang, Xinjiang and Ke, Zhanghan and Zhang, Wayne and Lau, Rynson WH},
  booktitle={CVPR},
  year={2023}
}

@inproceedings{convnext,
  title={A convnet for the 2020s},
  author={Liu, Zhuang and Mao, Hanzi and Wu, Chao-Yuan and Feichtenhofer, Christoph and Darrell, Trevor and Xie, Saining},
  booktitle={CVPR},
  year={2022}
}

@inproceedings{demystify_mamba,
  title={Demystify Mamba in Vision: A Linear Attention Perspective},
  author={Han, Dongchen and Wang, Ziyi and Xia, Zhuofan and Han, Yizeng and Pu, Yifan and Ge, Chunjiang and Song, Jun and Song, Shiji and Zheng, Bo and Huang, Gao},
  booktitle={NeurIPS},
  year={2024},
}

@inproceedings{dit,
  title={Scalable Diffusion Models with Transformers}, 
  author={William Peebles and Saining Xie},
  booktitle={ICCV},
  year={2023}
}

@inproceedings{shallow_vs_deep,
  title={Shallow vs. deep sum-product networks},
  author={Delalleau, Olivier and Bengio, Yoshua},
  booktitle={NeurIPS},
  year={2011}
}

@inproceedings{number_of_linear,
  title={On the number of linear regions of deep neural networks},
  author={Montufar, Guido F and Pascanu, Razvan and Cho, Kyunghyun and Bengio, Yoshua},
  booktitle={NeurIPS},
  year={2014}
}

@inproceedings{exponential_expressivity,
  title={Exponential expressivity in deep neural networks through transient chaos}, 
  author={Ben Poole and Subhaneil Lahiri and Maithra Raghu and Jascha Sohl-Dickstein and Surya Ganguli},
  booktitle={NeurIPS},
  year={2016}, 
}

@inproceedings{depth_power,
  title={The power of depth for feedforward neural networks},
  author={Eldan, Ronen and Shamir, Ohad},
  booktitle={Conference on learning theory},
  year={2016},
}

@inproceedings{benefits_of_depth,
  title={Benefits of depth in neural networks},
  author={Telgarsky, Matus},
  booktitle={Conference on learning theory},
  year={2016},
}

@inproceedings{expressive_power,
  title={On the Expressive Power of Deep Neural Networks}, 
  author={Maithra Raghu and Ben Poole and Jon Kleinberg and Surya Ganguli and Jascha Sohl-Dickstein},
  booktitle={ICML},      
  year={2017},
}

@inproceedings{counting_linear_regions,
  title={Bounding and counting linear regions of deep neural networks},
  author={Serra, Thiago and Tjandraatmadja, Christian and Ramalingam, Srikumar},
  booktitle={ICML},
  year={2018},
}

@inproceedings{resnet,
  title={Deep residual learning for image recognition},
  author={He, Kaiming and Zhang, Xiangyu and Ren, Shaoqing and Sun, Jian},
  booktitle={CVPR},
  year={2016}
}

@inproceedings{fractalnet,
  title={Fractalnet: Ultra-deep neural networks without residuals},
  author={Larsson, Gustav and Maire, Michael and Shakhnarovich, Gregory},
  booktitle={ICLR},
  year={2017}
}

@inproceedings{densenet,
  title={Densely connected convolutional networks},
  author={Huang, Gao and Liu, Zhuang and Van Der Maaten, Laurens and Weinberger, Kilian Q},
  booktitle={CVPR},
  year={2017}
}

@inproceedings{rezero,
  title={Rezero is all you need: Fast convergence at large depth},
  author={Bachlechner, Thomas and Majumder, Bodhisattwa Prasad and Mao, Henry and Cottrell, Gary and McAuley, Julian},
  booktitle={Uncertainty in Artificial Intelligence},
  year={2021},
}

@inproceedings{fixup,
  title={Fixup initialization: Residual learning without normalization},
  author={Zhang, Hongyi and Dauphin, Yann N and Ma, Tengyu},
  booktitle={ICLR},
  year={2019}
}

@inproceedings{layerscale,
  title={Going deeper with image transformers},
  author={Touvron, Hugo and Cord, Matthieu and Sablayrolles, Alexandre and Synnaeve, Gabriel and J{\'e}gou, Herv{\'e}},
  booktitle={ICCV},
  year={2021}
}

@article{deepnet,
  title={Deepnet: Scaling transformers to 1,000 layers},
  author={Wang, Hongyu and Ma, Shuming and Dong, Li and Huang, Shaohan and Zhang, Dongdong and Wei, Furu},
  journal={TPAMI},
  year={2024},
}

@article{normformer,
  title={Normformer: Improved transformer pretraining with extra normalization},
  author={Shleifer, Sam and Weston, Jason and Ott, Myle},
  journal={arXiv preprint arXiv:2110.09456},
  year={2021}
}

@article{deepvit,
  title={Deepvit: Towards deeper vision transformer},
  author={Zhou, Daquan and Kang, Bingyi and Jin, Xiaojie and Yang, Linjie and Lian, Xiaochen and Jiang, Zihang and Hou, Qibin and Feng, Jiashi},
  journal={arXiv preprint arXiv:2103.11886},
  year={2021}
}

@inproceedings{start_train,
  title={How to start training: The effect of initialization and architecture},
  author={Hanin, Boris and Rolnick, David},
  booktitle={NeurIPS},
  year={2018}
}

@article{universal_approx,
  title={Universal function approximation by deep neural nets with bounded width and relu activations},
  author={Hanin, Boris},
  journal={Mathematics},
  year={2019},
}

@inproceedings{min_width,
  title={Minimum width for universal approximation},
  author={Park, Sejun and Yun, Chulhee and Lee, Jaeho and Shin, Jinwoo},
  booktitle={ICLR},
  year={2021}
}

@inproceedings{min_width_2,
  title={Achieve the minimum width of neural networks for universal approximation},
  author={Cai, Yongqiang},
  booktitle={ICLR},
  year={2023}
}

@inproceedings{skipinit,
  title={Batch normalization biases residual blocks towards the identity function in deep networks},
  author={De, Soham and Smith, Sam},
  booktitle={NeurIPS},
  year={2020}
}

@article{llama,
  title={Llama: Open and efficient foundation language models},
  author={Touvron, Hugo and Lavril, Thibaut and Izacard, Gautier and Martinet, Xavier and Lachaux, Marie-Anne and Lacroix, Timoth{\'e}e and Rozi{\`e}re, Baptiste and Goyal, Naman and Hambro, Eric and Azhar, Faisal and others},
  journal={arXiv preprint arXiv:2302.13971},
  year={2023}
}

@inproceedings{wiki,
  title={Pointer sentinel mixture models},
  author={Merity, Stephen and Xiong, Caiming and Bradbury, James and Socher, Richard},
  booktitle={ICLR},
  year={2017}
}

@inproceedings{physics,
  title={Physics of language models: Part 2.1, grade-school math and the hidden reasoning process},
  author={Ye, Tian and Xu, Zicheng and Li, Yuanzhi and Allen-Zhu, Zeyuan},
  booktitle={ICLR},
  year={2025}
}

@inproceedings{agent_attention,
  title={Agent attention: On the integration of softmax and linear attention},
  author={Han, Dongchen and Ye, Tianzhu and Han, Yizeng and Xia, Zhuofan and Song, Shiji and Huang, Gao},
  booktitle={ECCV},
  year={2024},
}

@inproceedings{inline,
  title={Bridging the divide: Reconsidering softmax and linear attention},
  author={Han, Dongchen and Pu, Yifan and Xia, Zhuofan and Han, Yizeng and Pan, Xuran and Li, Xiu and Lu, Jiwen and Song, Shiji and Huang, Gao},
  booktitle={NeurIPS},
  year={2024}
}

@article{wang2025emulating,
  title={Emulating human-like adaptive vision for efficient and flexible machine visual perception},
  author={Wang, Yulin and Yue, Yang and Yue, Yang and Wang, Huanqian and Jiang, Haojun and Han, Yizeng and Ni, Zanlin and Pu, Yifan and Shi, Minglei and Lu, Rui and others},
  journal={Nature Machine Intelligence},
  year={2025},
}
}

\clearpage
\setcounter{page}{1}
\maketitlesupplementary

\section*{A. Datasets and Experiment Details}

\noindent
\textbf{ImageNet classification.} The ImageNet-1K~\cite{imagenet} dataset consists of 1.28M training images and 50K validation images. For a fair comparison, we train all our models under the same settings as the baseline models~\cite{deit, swin}. Specifically, our models are trained from scratch for 300 epochs using the AdamW~\cite{adamw} optimizer. We employ a cosine learning rate decay with a 20-epoch linear warm-up and a weight decay of 0.05. The base learning rate is set to $1 \times 10^{-3}$ for a batch size of 1024 and scaled linearly with batch size. Augmentation strategies include RandAugment~\cite{randaugment}, Mixup~\cite{mixup}, CutMix~\cite{cutmix}, and random erasing~\cite{random_erasing}.

\noindent
\textbf{COCO object detection.} The COCO dataset~\cite{coco}, with 118K training and 5K validation images, is a common benchmark for object detection and instance segmentation. We follow the standard Mask R-CNN~\cite{mrcn} and Cascade Mask R-CNN~\cite{crcnn} training settings to conduct experiments.

\noindent
\textbf{ADE20K semantic segmentation.} ADE20K \cite{ade20k} is a well-established benchmark for semantic segmentation which encompasses 20K training images, 2K validation images and 150 semantic categories. UPerNet~\cite{upernet} is used as the segmentation framework and the same training setting as Swin Transformer~\cite{swin} is adopted.

\noindent
\textbf{Language modeling.} WikiText-103~\cite{wiki} is a standard text dataset used for word-level language modeling. It contains 103M tokens from English Wikipedia articles. We train models on WikiText-103 with a vocabulary of size 50K, sequence length of 128, and batch size of 0.128M tokens.

\section*{B. Model Architectures}

We provide the architectures of ResNet~\cite{resnet}, DeiT~\cite{deit}, Swin~\cite{swin} and our Steps-ResNet, Steps-DeiT, Steps-Swin in \cref{tab:arch_resnet}, \cref{tab:arch_deit} and \cref{tab:arch_swin}. We simply apply the macro StepsNet architecture to these models while keeping their micro block design, patch embedding layers, and other network components unchanged. The macro design of Steps-Swin++ is the same as Steps-Swin, while its micro block employs components commonly used in advanced models~\cite{cmt, biformer} like LPU~\cite{cmt} and ConvFFN~\cite{cmt}.

\section*{C. Additional Results}

\noindent
\textbf{StepsNet at different stages.} 
For hierarchical models like Swin, we additionally investigate applying StepsNet at different stages. As depicted in \cref{tab:ablation_stage}, substituting StepsNet in the last two stages improves performance, while changes in the first two stages yield minimal effect. This could be attributed to the larger widths of last two stages, which support relatively wide sub-networks and enable StepsNet to better benefit the model with increased depth.

\begin{table}[h]
    \centering
    \footnotesize
    \setlength{\tabcolsep}{0.5mm}{
    \renewcommand\arraystretch{1.05}
    \begin{tabular}{c c c c|c c|c c|c}
        \bottomrule
        \multicolumn{4}{c|}{Stages w/ StepsNet} & \multirow{2}{*}{\#Blocks} & \multirow{2}{*}{\#Layers} & \multirow{2}{*}{\#Param} & \multirow{2}{*}{FLOPS} & \multirow{2}{*}{Acc.} \\
        Stage1      & Stage2        & Stage3        & Stage4        &       &       &       &       &       \\
        \hline
        $\checkmark$&               &               &               & 13    & 70    & 28.3M & 4.5G  & 81.3  \\
                    &$\checkmark$   &               &               & 13    & 70    & 28.3M & 4.5G  & 81.3  \\
                    &               &$\checkmark$   &               & 18    & 95    & 28.3M & 4.5G  & 81.5  \\
                    &               &               &$\checkmark$   & 20    & 105   & 27.8M & 4.5G  & 82.2  \\
        \rowcolor{lightgray!20}
                    &               &$\checkmark$   &$\checkmark$   & 26    & 135   & 27.8M & 4.5G  & 82.4  \\
        \hline
        \multicolumn{4}{c|}{Swin-T}                                 & 12    & 65    & 28.3M & 4.5G  & 81.3  \\
        \toprule
    \end{tabular}}
    \vskip -0.2cm
    \caption{Applying StepsNet at different stages of Swin-T model.}
    \label{tab:ablation_stage}
    \vskip -0.3cm
\end{table}

\begin{table*}[h]
\centering
\scriptsize
\setlength{\tabcolsep}{1.5mm}{
\renewcommand\arraystretch{1.2}
    \begin{tabular}{c|c|c|c|c}
    \toprule
    \textbf{Model} & \textbf{Stage 1} & \textbf{Stage 2} & \textbf{Stage 3} & \textbf{Stage 4} \\
    \midrule
    ResNet-18~\cite{resnet}
    & $64 \!\times\! 1$ 
    & $128 \!\times\! 1$
    & $256 \!\times\! 1$
    & $512 \!\times\! 1$ \\

    \midrule

    ResNet-34~\cite{resnet}
    & $64 \!\times\! 2$ 
    & $128 \!\times\! 3$
    & $256 \!\times\! 5$
    & $512 \!\times\! 2$ \\

    \midrule

    ResNet-50~\cite{resnet}
    & $64 \!\times\! 2$ 
    & $128 \!\times\! 3$
    & $256 \!\times\! 5$
    & $512 \!\times\! 2$ \\
    
    \midrule
    
    \cellcolor{lightgray!20} Steps-ResNet-18
    & $64 \!\times\! 1$ 
    & $128 \!\times\! 1$
    & $64 \!\times\! 2,\ 90 \!\times\! 1,\ 128 \!\times\! 1,\ 181 \!\times\! 1,\ 256 \!\times\! 0$
    & $64 \!\times\! 2,\ 90 \!\times\! 1,\ 128 \!\times\! 1,\ 181 \!\times\! 1,\ 256 \!\times\! 1,\ 362 \!\times\! 1,\ 512 \!\times\! 0$ \\

    \midrule

    \cellcolor{lightgray!20} Steps-ResNet-34
    & $64 \!\times\! 1$ 
    & $128 \!\times\! 1$
    & $64 \!\times\! 6,\ 90 \!\times\! 3,\ 128 \!\times\! 3,\ 181 \!\times\! 3,\ 256 \!\times\! 2$
    & $64 \!\times\! 2,\ 90 \!\times\! 1,\ 128 \!\times\! 1,\ 181 \!\times\! 1,\ 256 \!\times\! 1,\ 362 \!\times\! 1,\ 512 \!\times\! 1$ \\

    \midrule

    \cellcolor{lightgray!20} Steps-ResNet-50
    & $64 \!\times\! 1$ 
    & $128 \!\times\! 1$
    & $64 \!\times\! 6,\ 90 \!\times\! 3,\ 128 \!\times\! 3,\ 181 \!\times\! 3,\ 256 \!\times\! 2$
    & $64 \!\times\! 2,\ 90 \!\times\! 1,\ 128 \!\times\! 1,\ 181 \!\times\! 1,\ 256 \!\times\! 1,\ 362 \!\times\! 1,\ 512 \!\times\! 1$ \\
    \bottomrule
    \end{tabular}}
\caption{Architectures of ResNet~\cite{resnet} and Steps-ResNet models of the form $\rm{C}\times\rm{D}$, where $\rm{C}$ is the bottleneck dimension and $\rm{D}$ denotes number of blocks. Notably, in ResNet, the first block of each stage is a downsample block, which is not listed in this table.}
\label{tab:arch_resnet}
\end{table*}

\begin{table*}[h]
\centering
\scriptsize
\setlength{\tabcolsep}{1.5mm}{
\renewcommand\arraystretch{1.2}
    \begin{tabular}{c|c}
    \toprule
    \textbf{Model} & \textbf{Architecture} \\
    \midrule
    DeiT-T~\cite{deit}
    & $
    \left[\!\!\! \begin{array}{c} {\rm C} \ 192 \\ {\rm H} \ 3 \end{array} \!\!\! \right ] \!\!\times\! 12$ \\

    \midrule

    DeiT-S~\cite{deit}
    & $
    \left[\!\!\! \begin{array}{c} {\rm C} \ 384 \\ {\rm H} \ 6 \end{array} \!\!\! \right ] \!\!\times\! 12$ \\

    \midrule

    DeiT-B~\cite{deit}
    & $
    \left[\!\!\! \begin{array}{c} {\rm C} \ 768 \\ {\rm H} \ 12 \end{array} \!\!\! \right ] \!\!\times\! 12$ \\

    \midrule
    
    \cellcolor{lightgray!20} Steps-DeiT-T
    & $
    \left[\!\!\! \begin{array}{c} {\rm C} \ 96 \\ {\rm H} \ 2 \end{array} \!\!\! \right ] \!\!\times\! 12,
    \left[\!\!\! \begin{array}{c} {\rm C} \ 136 \\ {\rm H} \ 2 \end{array} \!\!\! \right ] \!\!\times\! 6,
    \left[\!\!\! \begin{array}{c} {\rm C} \ 192 \\ {\rm H} \ 3 \end{array} \!\!\! \right ] \!\!\times\! 6$ \\

    \midrule

    \cellcolor{lightgray!20} Steps-DeiT-S
    & $
    \left[\!\!\! \begin{array}{c} {\rm C} \ 192 \\ {\rm H} \ 3 \end{array} \!\!\! \right ] \!\!\times\! 12,
    \left[\!\!\! \begin{array}{c} {\rm C} \ 272 \\ {\rm H} \ 4 \end{array} \!\!\! \right ] \!\!\times\! 6,
    \left[\!\!\! \begin{array}{c} {\rm C} \ 384 \\ {\rm H} \ 6 \end{array} \!\!\! \right ] \!\!\times\! 6$ \\

    \midrule

    \cellcolor{lightgray!20} Steps-DeiT-B
    & $
    \left[\!\!\! \begin{array}{c} {\rm C} \ 96 \\ {\rm H} \ 2 \end{array} \!\!\! \right ] \!\!\times\! 12,
    \left[\!\!\! \begin{array}{c} {\rm C} \ 136 \\ {\rm H} \ 2 \end{array} \!\!\! \right ] \!\!\times\! 12,
    \left[\!\!\! \begin{array}{c} {\rm C} \ 192 \\ {\rm H} \ 3 \end{array} \!\!\! \right ] \!\!\times\! 12,
    \left[\!\!\! \begin{array}{c} {\rm C} \ 272 \\ {\rm H} \ 4 \end{array} \!\!\! \right ] \!\!\times\! 12,
    \left[\!\!\! \begin{array}{c} {\rm C} \ 384 \\ {\rm H} \ 6 \end{array} \!\!\! \right ] \!\!\times\! 12$ \\
    \bottomrule
    \end{tabular}}
\caption{Architectures of DeiT~\cite{deit} and Steps-DeiT models. C and H denote block width and number of heads, respectively.}
\label{tab:arch_deit}
\end{table*}

\begin{table*}[h]
\centering
\scriptsize
\setlength{\tabcolsep}{0.5mm}{
\renewcommand\arraystretch{1.2}
    \begin{tabular}{c|c|c|c|c}
    \toprule
    \textbf{Model} & \textbf{Stage 1} & \textbf{Stage 2} & \textbf{Stage 3} & \textbf{Stage 4} \\
    \midrule
    Swin-T~\cite{swin}
    & $\left[\!\!\! \begin{array}{c} {\rm C} \ 96 \\ {\rm H} \ 3 \end{array} \!\!\! \right ] \!\!\times\! 2$ 
    & $\left[\!\!\! \begin{array}{c} {\rm C} \ 192 \\ {\rm H} \ 6 \end{array} \!\!\! \right ] \!\!\times\! 2$
    & $\left[\!\!\! \begin{array}{c} {\rm C} \ 384 \\ {\rm H} \ 12 \end{array} \!\!\! \right ] \!\!\times\! 6$
    & $\left[\!\!\! \begin{array}{c} {\rm C} \ 768 \\ {\rm H} \ 24 \end{array} \!\!\! \right ] \!\!\times\! 2$ \\

    \midrule

    Swin-S~\cite{swin}
    & $\left[\!\!\! \begin{array}{c} {\rm C} \ 96 \\ {\rm H} \ 3 \end{array} \!\!\! \right ] \!\!\times\! 2$ 
    & $\left[\!\!\! \begin{array}{c} {\rm C} \ 192 \\ {\rm H} \ 6 \end{array} \!\!\! \right ] \!\!\times\! 2$
    & $\left[\!\!\! \begin{array}{c} {\rm C} \ 384 \\ {\rm H} \ 12 \end{array} \!\!\! \right ] \!\!\times\! 18$
    & $\left[\!\!\! \begin{array}{c} {\rm C} \ 768 \\ {\rm H} \ 24 \end{array} \!\!\! \right ] \!\!\times\! 2$ \\

    \midrule

    Swin-B~\cite{swin}
    & $\left[\!\!\! \begin{array}{c} {\rm C} \ 128 \\ {\rm H} \ 4 \end{array} \!\!\! \right ] \!\!\times\! 2$ 
    & $\left[\!\!\! \begin{array}{c} {\rm C} \ 256 \\ {\rm H} \ 8 \end{array} \!\!\! \right ] \!\!\times\! 2$
    & $\left[\!\!\! \begin{array}{c} {\rm C} \ 512 \\ {\rm H} \ 16 \end{array} \!\!\! \right ] \!\!\times\! 18$
    & $\left[\!\!\! \begin{array}{c} {\rm C} \ 1024 \\ {\rm H} \ 32 \end{array} \!\!\! \right ] \!\!\times\! 2$ \\
    
    \midrule
    
    \cellcolor{lightgray!20} Steps-Swin-T
    & $\left[\!\!\! \begin{array}{c} {\rm C} \ 96 \\ {\rm H} \ 3 \end{array} \!\!\! \right ] \!\!\times\! 2$ 
    & $\left[\!\!\! \begin{array}{c} {\rm C} \ 192 \\ {\rm H} \ 6 \end{array} \!\!\! \right ] \!\!\times\! 2$
    & $\left[\!\!\! \begin{array}{c} {\rm C} \ 192 \\ {\rm H} \ 6 \end{array} \!\!\! \right ] \!\!\times\! 6,
    \left[\!\!\! \begin{array}{c} {\rm C} \ 272 \\ {\rm H} \ 8 \end{array} \!\!\! \right ] \!\!\times\! 3,
    \left[\!\!\! \begin{array}{c} {\rm C} \ 384 \\ {\rm H} \ 12 \end{array} \!\!\! \right ] \!\!\times\! 3$
    & $\left[\!\!\! \begin{array}{c} {\rm C} \ 192 \\ {\rm H} \ 6 \end{array} \!\!\! \right ] \!\!\times\! 4,
    \left[\!\!\! \begin{array}{c} {\rm C} \ 272 \\ {\rm H} \ 8 \end{array} \!\!\! \right ] \!\!\times\! 2,
    \left[\!\!\! \begin{array}{c} {\rm C} \ 384 \\ {\rm H} \ 12 \end{array} \!\!\! \right ] \!\!\times\! 2,
    \left[\!\!\! \begin{array}{c} {\rm C} \ 544 \\ {\rm H} \ 16 \end{array} \!\!\! \right ] \!\!\times\! 2,
    \left[\!\!\! \begin{array}{c} {\rm C} \ 768 \\ {\rm H} \ 24 \end{array} \!\!\! \right ] \!\!\times\! 0$ \\

    \midrule

    \cellcolor{lightgray!20} Steps-Swin-S
    & $\left[\!\!\! \begin{array}{c} {\rm C} \ 96 \\ {\rm H} \ 3 \end{array} \!\!\! \right ] \!\!\times\! 2$ 
    & $\left[\!\!\! \begin{array}{c} {\rm C} \ 192 \\ {\rm H} \ 6 \end{array} \!\!\! \right ] \!\!\times\! 2$
    & $\left[\!\!\! \begin{array}{c} {\rm C} \ 192 \\ {\rm H} \ 6 \end{array} \!\!\! \right ] \!\!\times\! 6,
    \left[\!\!\! \begin{array}{c} {\rm C} \ 272 \\ {\rm H} \ 8 \end{array} \!\!\! \right ] \!\!\times\! 3,
    \left[\!\!\! \begin{array}{c} {\rm C} \ 384 \\ {\rm H} \ 12 \end{array} \!\!\! \right ] \!\!\times\! 15$
    & $\left[\!\!\! \begin{array}{c} {\rm C} \ 192 \\ {\rm H} \ 6 \end{array} \!\!\! \right ] \!\!\times\! 4,
    \left[\!\!\! \begin{array}{c} {\rm C} \ 272 \\ {\rm H} \ 8 \end{array} \!\!\! \right ] \!\!\times\! 2,
    \left[\!\!\! \begin{array}{c} {\rm C} \ 384 \\ {\rm H} \ 12 \end{array} \!\!\! \right ] \!\!\times\! 2,
    \left[\!\!\! \begin{array}{c} {\rm C} \ 544 \\ {\rm H} \ 16 \end{array} \!\!\! \right ] \!\!\times\! 2,
    \left[\!\!\! \begin{array}{c} {\rm C} \ 768 \\ {\rm H} \ 24 \end{array} \!\!\! \right ] \!\!\times\! 0$ \\

    \midrule

    \cellcolor{lightgray!20} Steps-Swin-B
    & $\left[\!\!\! \begin{array}{c} {\rm C} \ 128 \\ {\rm H} \ 4 \end{array} \!\!\! \right ] \!\!\times\! 2$ 
    & $\left[\!\!\! \begin{array}{c} {\rm C} \ 256 \\ {\rm H} \ 8 \end{array} \!\!\! \right ] \!\!\times\! 2$
    & $\left[\!\!\! \begin{array}{c} {\rm C} \ 256 \\ {\rm H} \ 8 \end{array} \!\!\! \right ] \!\!\times\! 6,
    \left[\!\!\! \begin{array}{c} {\rm C} \ 360 \\ {\rm H} \ 12 \end{array} \!\!\! \right ] \!\!\times\! 3,
    \left[\!\!\! \begin{array}{c} {\rm C} \ 512 \\ {\rm H} \ 16 \end{array} \!\!\! \right ] \!\!\times\! 15$
    & $\left[\!\!\! \begin{array}{c} {\rm C} \ 256 \\ {\rm H} \ 8 \end{array} \!\!\! \right ] \!\!\times\! 4,
    \left[\!\!\! \begin{array}{c} {\rm C} \ 360 \\ {\rm H} \ 12 \end{array} \!\!\! \right ] \!\!\times\! 2,
    \left[\!\!\! \begin{array}{c} {\rm C} \ 512 \\ {\rm H} \ 16 \end{array} \!\!\! \right ] \!\!\times\! 2,
    \left[\!\!\! \begin{array}{c} {\rm C} \ 720 \\ {\rm H} \ 24 \end{array} \!\!\! \right ] \!\!\times\! 2,
    \left[\!\!\! \begin{array}{c} {\rm C} \ 1024 \\ {\rm H} \ 32 \end{array} \!\!\! \right ] \!\!\times\! 0$ \\
    \bottomrule
    \end{tabular}}
\caption{Architectures of Swin~\cite{swin} and Steps-Swin models. C and H denote block width and number of heads, respectively.}
\label{tab:arch_swin}
\end{table*}


\end{document}